\documentclass[10pt,twocolumn,letterpaper]{article}

\usepackage{cvpr}
\usepackage{times}
\usepackage[dvipdfmx]{graphicx}
\usepackage{amsmath}
\usepackage{amssymb}
\usepackage{bm}
\usepackage{braket}

\usepackage[breaklinks=true,bookmarks=false]{hyperref}

\cvprfinalcopy 

\def\cvprPaperID{****} 

\begin{document}

\title{SwGridNet: A Deep Convolutional Neural Network based on Grid Topology\\ for Image Classification}

\author{Atsushi Takeda\\
Tohoku Gakuin University\\
2-1-1, Tenjinzawa, Izumi-ku, Sendai, Japan\\
{\tt\small takeda@cs.tohoku-gakuin.ac.jp}
}

\maketitle

\begin{abstract}
Deep convolutional neural networks (CNNs) achieve remarkable performance on image classification tasks.
Recent studies, however, have demonstrated that generalization abilities are more important than the depth of neural networks for improving performance on image classification tasks.
Herein, a new neural network called SwGridNet is proposed.
A SwGridNet includes many convolutional processing units which connect mutually as a grid network where many processing paths exist between input and output.
A SwGridNet has high generalization capability because the multipath architecture has the same effect of ensemble learning.
As described in this paper, details of the SwGridNet network architecture are presented.
Experimentally obtained results presented in this paper show that SwGridNets respectively achieve test error rates of 2.95\% and 15.67\% in a CIFAR-10 and CIFAR-100 classification tasks.
The results indicate that the SwGridNet performance approximates that of state-of-the-art deep CNNs.
\end{abstract}

\section{Introduction}
\label{sec:introduction}

A deep convolutional neural network AlexNet~\cite{AlexNet} won the ImageNet Large Scale Visual Recognition Competition (ILSVRC)~\cite{ILSVRC} in 2012.
This result demonstrated that deep convolutional neural networks (CNNs) are suitable for image recognition tasks.
Therefore, in recent years, many researchers have been investigating deep CNNs for use in image recognition.
Although it is not easy to train deep neural networks,
deep neural networks generally have higher capabilities than shallow neural networks.
Therefore, many techniques have been proposed for training deep CNNs.

Deep CNNs achieve high accuracy rates for image classification tasks~\cite{ResNet,WideResNet,Xception,PyramidalResNet,ResNeXt}.
Deep CNNs, however, tend to memorize all training data. For that reason, a mechanism against overfitting is necessary for deep CNNs~\cite{UnderstandingDeepLearning}.
A lot of successful deep CNNs such as Xception~\cite{Xception} or ResNeXt~\cite{ResNeXt} are constructed as a multipath network
which has many processing paths between input and output.
In a multipath network, each processing path performs the calculation using individual parameters.
The results are integrated as output data.
A multipath network mitigates overfitting and improves generalization abilities
because a multipath network has the same effect as ensemble learning.

As described in this paper, a new deep convolutional neural network called {\it Sandwiched Grid convolutional neural Network (SwGridNet)} is proposed.
A SwGridNet is constructed as a chain of grid blocks which include many convolutional processing units.
The convolutional processing unit mutually connects as a grid network with many processing paths.
Because a SwGridNet contains many processing paths between input and output,
the SwGridNet gains high generalization capabilities.
This paper presents an explanation of SwGridNet architecture
and describes that a SwGridNet has many processing paths between input and output.
In addition, experimentally obtained results of image classification tasks of CIFAR-10 and CIFAR-100 dataset~\cite{CIFAR} are presented.
The experiment results show that SwGridNets yield test error rates of 2.95\% and 15.67\%, respectively, for CIFAR-10 and CIFAR100 classification tasks.

The organization of this paper is the following.
Section~\ref{sec:related} introduces existing deep CNNs, which have been proposed for image classification tasks, and presents explanation importance of generalization abilities of deep CNNs.
Section~\ref{sec:proposal} presents a proposal of a new deep convolutional neural network called SwGridNet.
This section explains details of a SwGridNet architecture,
and explains that a SwGridNet contains many processing paths between input and output.
Section~\ref{sec:evaluation} presents experimentally obtained results of SwGridNets in CIFAR-10 and CIFAR100 classification tasks.
The experimentally obtained results show that SwGridNets achieve low error rates which are close to other state-of-the-art deep CNNs.
Finally, the study is concluded in section~\ref{sec:conclusion}.

\section{Related Works}
\label{sec:related}

\subsection{Deep CNN for Image Classification}
A deep convolutional neural network (CNN) is the most successful method for image recognition tasks.
For that reason, many types of deep CNNs for image classification tasks have been proposed.
AlexNet~\cite{AlexNet} and VGGNet~\cite{VGGNet} respectively achieved outstanding results in ILSVRC 2012 and 2014.
Results showed that deep CNNs work well in image classification tasks.
Deep CNNs generally have higher capabilities than shallow CNNs, but training a deep CNN is not easy because of vanishing gradient problems~\cite{VanishingGradient1,VanishingGradient2}.

To avoid vanishing gradient problems, state-of-the-art deep CNNs contain not only convolutional layers but also shortcuts.
A ResNet is constructed as a chain of Residual Blocks which include both convolutional layers and a shortcut~\cite{ResNet}.
The shortcut leads to an error signal directly from output to input.
For that reason, it is possible to train a very deep ResNet which includes more than 1,000 convolutional layers.
Other state-of-the-art deep CNNs such as Xception~\cite{Xception}, FractalNet~\cite{FractalNet} and ResNeXt~\cite{ResNeXt}
also have shortcuts between input and output.

\subsection{Generalization Ability of Deep CNN}

\begin{figure}
\centering
 \includegraphics[width=\hsize]{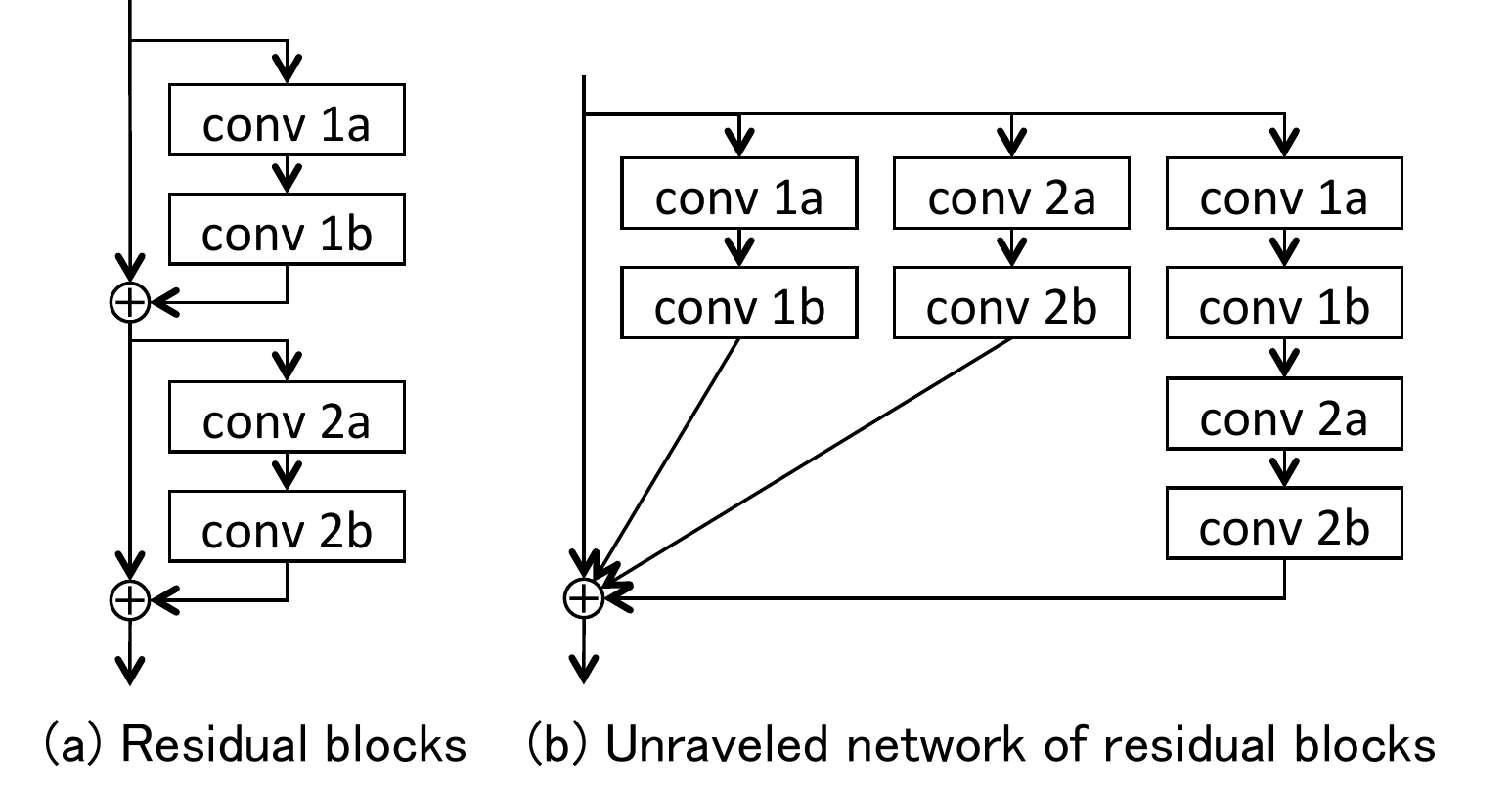}
 \caption{Residual networks that include two residual blocks (four convolutional layers): networks presented in (a) and (b) are the same convolutional neural networks.}
\label{fig:resnet}
\end{figure}

Deep CNNs have numerous parameters, so deep CNNs can memorize much information.
In fact, AlexNet and Inception can perfectly memorize CIFAR-10 training images that are labeled randomly~\cite{UnderstandingDeepLearning}.
Therefore, deep CNNs easily overfit the training data.
To improve test accuracy rates, generalization abilities are required for deep CNNs.
In recent years, many generalization techniques such as Dropout~\cite{Dropout} and Shakeout~\cite{Shakeout} have been proposed.
In addition, many state-of-the-art deep CNNs are constructed as a multipath neural network
that contains more than one processing path to improve the generalization ability.
In a multipath neural network, each processing path performs the calculation using each parameter.
The network integrates the results from all processing paths as output data.
A multipath neural network has the same effect as ensemble learning~\cite{EnsembleLearning} which improves the capability of a neural network for generalization.
Therefore, a multipath architecture is used in many state-of-the-art deep CNNs.

Inception~\cite{Inception} and Xception~\cite{Xception} networks are constructed as a chain of Inception modules
that contain more than one convolutional layer connected in parallel.
Therefore, Inception and Xception have many processing paths between input and output.
A FractalNet~\cite{FractalNet} is also a multipath neural network.
A FractalNet contains not only shallow processing paths but also deep processing paths.
A Residual Network (ResNet)~\cite{ResNet} is constructed as a chain of Residual Blocks
which are the same as multipath neural networks as shown in Figure~\ref{fig:resnet}.
Therefore, a ResNet has numerous processing paths between input and output.
Wide Residual Network~\cite{WideResNet}, Pyramidal Residual Network~\cite{PyramidalResNet} and ResNeXt~\cite{ResNeXt} are
extended versions of a ResNet. These networks are also multipath neural networks.
Because the multipath neural networks show remarkable performances in image classification tasks,
A multipath architecture is an important technique for use with deep CNNs.

As described herein, a new multipath convolutional neural network is proposed {\it Sandwiched Grid convolutional neural Network (SwGridNet)}.
A SwGridNet has many convolutional processing units which connect mutually as a grid network.
A grid network contains many processing paths between input and output.
Therefore, a SwGridNet is a multipath neural network which has a high generalization ability.
An existing convolutional neural network GridNet~\cite{GridNet} is also constructed using grid network topology.
A GridNet, however, uses a grid topology for the integration of different resolution feature maps.
However, a SwGridNet uses a grid topology for improvement of generalization capability.
\section{SwGridNet}
\label{sec:proposal}

\subsection{Architecture of SwGridNet}
\label{subsec:arch}

\begin{figure}
\centering
 \includegraphics[width=\hsize]{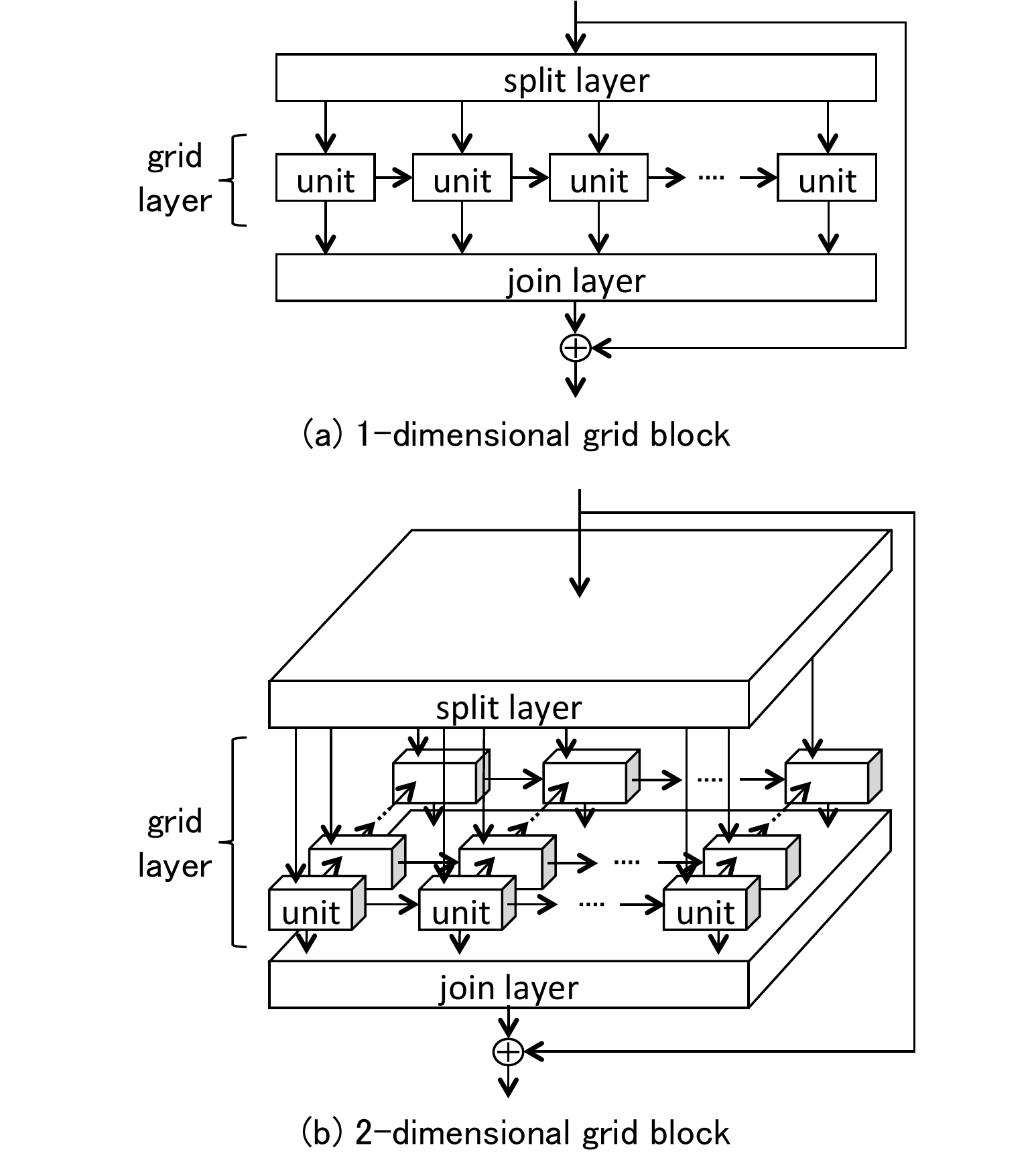}
 \caption{Architectures of grid blocks in a SwGridNet.}
 \label{fig:gridnet1}
\end{figure}

A SwGridNet is constructed as a chain of grid blocks.
A grid block includes a residual architecture to lead error signals directly from output to input.
A function of a grid block is defined as
\begin{eqnarray}
{\bm y} = \mathcal{F}({\bm x}) + {\bm x}.
\end{eqnarray}
Here, ${\bm x}$ and ${\bm y} $ respectively denote an input matrix and an output matrix of the grid block.
The function $\mathcal{F}(\cdot)$ represents a nonlinear function of the grid block.
A grid block includes a split layer, a grid layer, and a join layer, as shown in Figure~\ref{fig:gridnet1}.
The function $\mathcal{F}(\cdot)$ is separated to three functions as
\begin{eqnarray}
\mathcal{F}({\bm x}) := \mathcal{G}^{\textrm{join}}(\mathcal{G}^{\textrm{grid}}(\mathcal{G}^{\textrm{split}}({\bm x}))).
\end{eqnarray}
Here, $\mathcal{G}^{\textrm{split}}(\cdot)$, $\mathcal{G}^{\textrm{grid}}(\cdot)$ and $\mathcal{G}^{\textrm{join}}(\cdot)$ respectively denote nonlinear functions
of a split layer, a grid layer, and a join layer.

A grid layer contains convolutional processing units which connect mutually as a multi-dimensional grid network.
Consequently, the number of the units in a grid layer is $L^D$,
where $D$ denotes the dimension of the grid layer and $L$ denotes the side length of the grid layer.
As described in this paper, a location of the unit is represented as a coordinate $(p_0,p_1,\cdots,p_{N-1})$.
Each unit includes convolutional layers, which have individual parameters.

\begin{figure}
\centering
 \includegraphics[width=\hsize]{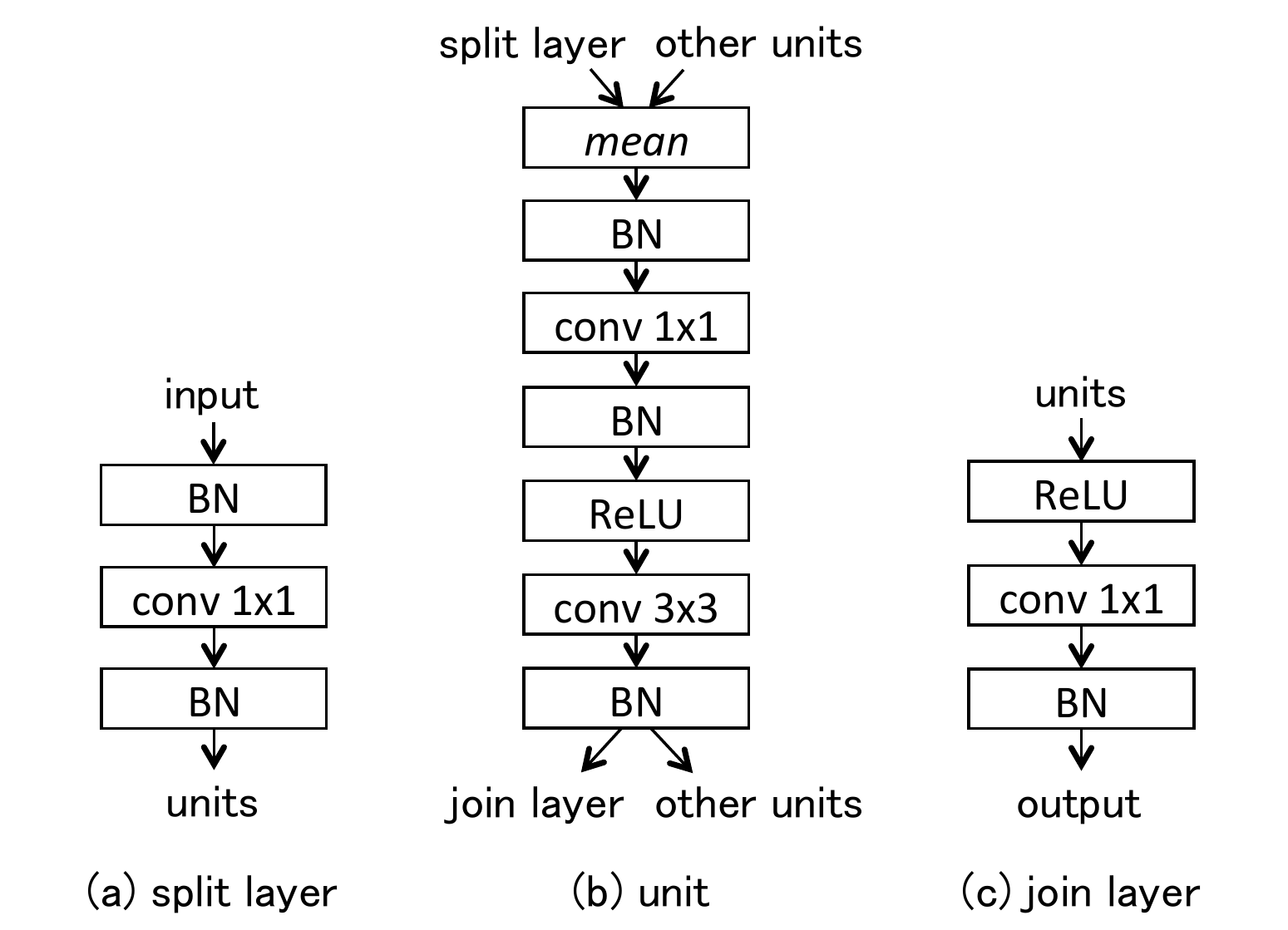}
 \caption{Calculations in a grid block:
(a) depicts a chain of calculations in a split layer,
(b) depicts a chain of calculations in a convolutional processing unit,
and (c) depicts a chain of calculations in a join layer.
In these chains, `conv', `BN', and `ReLU' respectively represent a convolutional layer, a batch normalization layer and a ReLU function.
In addition, `{\it mean}' is an element-wise mean calculation.}
 \label{fig:gridnet2}
\end{figure}

A split layer creates matrices input to all convolutional processing units at first.
The calculation of a split layer is defined as
\begin{eqnarray}
\set{{\bm s}_{p_0,\cdots,p_{N-1}} \mid 0 \leq p_i < L} = \mathcal{G}^{\textrm{split}}({\bm x}).
\end{eqnarray}
Here, ${\bm s}_{p_0,\cdots,p_{N-1}}$ denotes an output matrix that is sent to a unit located at $(p_0,\cdots,p_{N-1})$.
To perform the function $\mathcal{G}^{\textrm{split}}$, a split layer is constructed as a neural network that contains a convolution layer and batch normalization layers~\cite{BatchNormalization} as shown in Figure~\ref{fig:gridnet2}~(a).

The calculation of a grid layer is defined as
\begin{eqnarray}
\renewcommand{\arraystretch}{1.3}
\begin{array}{l}
 \set{{\bm u}_{p_0,\cdots,p_{N-1}} \mid 0 \leq p_i < L} \\
\quad\quad\quad\quad = \mathcal{G}^{\textrm{grid}}(\set{{\bf s}_{p_0,\cdots,p_{N-1}} \mid 0 \leq p_i < L}).
\end{array}
\renewcommand{\arraystretch}{1}
\end{eqnarray}
Here, ${\bm u}_{p_0,\cdots,p_{N-1}}$ is a set of output matrices of a grid layer.
To perform the function $\mathcal{G}^{\textrm{grid}}$, a grid layer contains many convolutional processing units, as shown in Figure~\ref{fig:gridnet1}.
I define a function performed by an convolutional processing unit which is located at $(p_0,\cdots,p_{N-1})$ in a grid layer as
\begin{eqnarray}
 {\bm u}_{p_0,\cdots,p_{N-1}} = \mathcal{H}({\bf s}_{p_0,\cdots,p_{N-1}}, {\bm v}_{p_0,\cdots,p_{N-1}}), \\
 {\bm v}_{p_0,\cdots,p_{N-1}} = \set{ \bm{u}_{p_0,\cdots,p_m-1,\cdots,p_{N-1}} \mid p_m > 0}.
\end{eqnarray}
Here, ${\bm v}_{p_0,\cdots,p_{N-1}}$ is a set of output matrices of neighbor units;
$\mathcal{H}(\cdot)$ represents a nonlinear function of the unit.
Each unit in a grid layer receives not only a matrix ${\bm s}_{p_0,\cdots,p_{N-1}}$ from a split layer but also a set of matrices ${\bm v}_{p_0,\cdots,p_{N-1}}$ from neighbor units.
To perform the function $\mathcal{H}(\cdot)$, each convolutional processing unit in a grid layer is also constructed as a neural network that contains convolution layers, batch normalization layers and a ReLU~\cite{ReLU} function as shown in Figure~\ref{fig:gridnet2}~(b).

A set of matrix output by a grid layer is sent to a join layer as input matrices.
Calculation of a join layer is defined as
\begin{eqnarray}
{\bm y} = \mathcal{G}^{\textrm{join}}(\set{{\bm u}_{p_0,\cdots,p_{N-1}} \mid 0 \leq p_i < L}).
\end{eqnarray}
To perform the function $\mathcal{G}^{\textrm{join}}$, a join layer is constructed as a neural network that contains a convolution layer, a batch normalization layer and a ReLU function, as portrayed in Figure~\ref{fig:gridnet2}~(c).

\subsection{Processing Paths in a Grid Block}
\label{subsec:path}

\begin{figure}
\centering
 \includegraphics[width=\hsize]{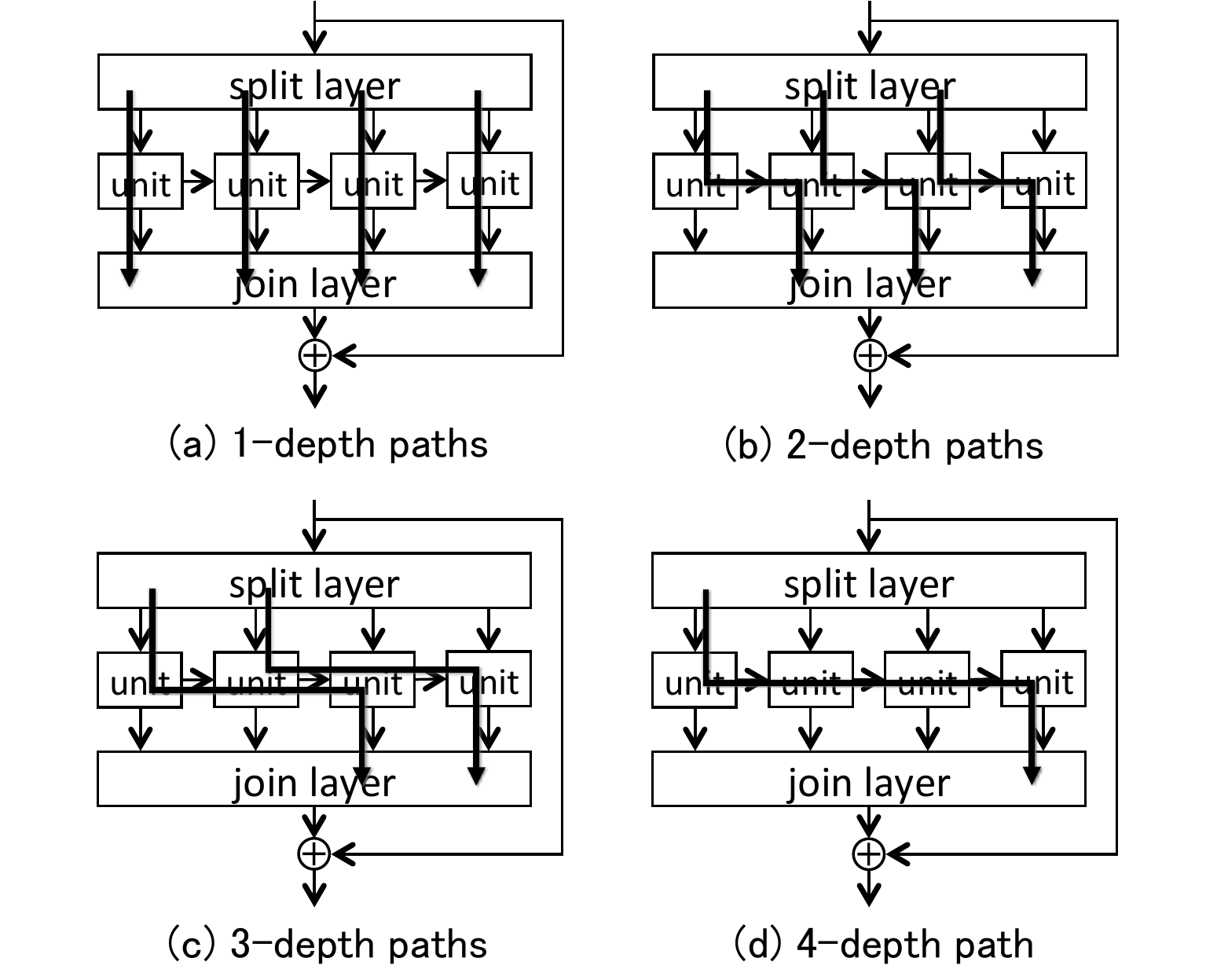}
 \caption{Processing paths in a one-dimensional grid block with four convolutional processing units.}
 \label{fig:gridnet3}
\end{figure}

A grid block contains many convolutional processing units, which are mutually connected as a grid network.
More than one communication path connects a source node to a destination node in a grid network.
In the same way as a grid network, a grid block also has more than one processing path between input and output.
Figure~\ref{fig:gridnet3} shows processing paths in a one-dimensional grid block that consists of four convolutional processing units.
This grid block shown in Figure~\ref{fig:gridnet3} has four depth processing paths.
Here, the processing path depth represents the number of units passed by the processing path.
In the case of a one-dimensional grid block with four convolutional processing units,
the grid block arranges four one-depth processing paths, as shown in Figure~\ref{fig:gridnet3}~(a).
In the same manner, the grid block also arranges three two-depth processing paths, two three-depth processing paths, and one four-depth path.
The number of processing paths depends on dimension $N$ and the side length $L$ of the grid block.
The deepest processing path in the grid block is an $N(L-1)$-depth path.
Each processing path in a grid block performs calculations using individual parameters,
so the multipath architecture of a grid block has the same effect as ensemble learning.

\begin{table}[t]
\centering
 \caption{Number of processing paths in a grid block.
 Here, $D$ and $L $ respectively denote the dimension and the side length of the grid block.}
 \label{table:path}
 \begin{tabular}{|l|c|c|c|}
  \hline
  & $N$=1, $L$=16 & $N$=2, $L$=4 & $N$=4, $L$=2 \\
  \hline
  depth=1 & 16 & 16 & 16 \\
  depth=2 & 15 & 24 & 32 \\
  depth=3 & 14 & 34 & 48 \\
  depth=4 & 13 & 44 & 48 \\
  depth=5 & 12 & 48 & 24 \\
  depth=6 & 11 & 40 & 0 \\
  depth=7 & 10 & 20 & 0 \\
  depth=8 & 9 & 0 & 0 \\
  depth=9 & 8 & 0 & 0 \\
  depth=10 & 7 & 0 & 0 \\
  depth=11 & 6 & 0 & 0 \\
  depth=12 & 5 & 0 & 0 \\
  depth=13 & 4 & 0 & 0 \\
  depth=14 & 3 & 0 & 0 \\
  depth=15 & 2 & 0 & 0 \\
  depth=15 & 1 & 0 & 0 \\
  \hline 
  total & 136 & 226 & 168 \\
  \hline 
 \end{tabular}
\end{table}

Table~\ref{table:path} presents the number of processing paths in a grid block that consists of 16 convolutional processing units.
A four-dimensional grid block ($N$=4, $L$=2) has more shallow processing paths than a one-dimensional grid block ($N$=1, $L$=16).
Therefore, the effect of ensemble learning in a four-dimensional grid block is better than a one-dimensional grid block
because the effect of ensemble learning is generally improved along with the increased number of processing paths.
However, a four-dimensional grid block has no deeper processing path than five-depth, but a one-dimensional grid block has a 16-depth processing path.
Therefore, a one-dimensional grid block can support more complex calculation than a four-dimensional grid block.
A report of a study of deep CNNs~\cite{ResNetBehave} indicates that ensemble learning by shallow processing paths is more effective than calculations of a deep processing path in image classification tasks.
In addition, many experiment results support the report~\cite{WideResNet}.
Therefore, a four-dimensional grid block is expected to be more suitable than a one-dimensional grid block for image classification tasks.

\subsection{Unit Width in a Grid Block}

\begin{figure}
\centering
 \includegraphics[width=\hsize]{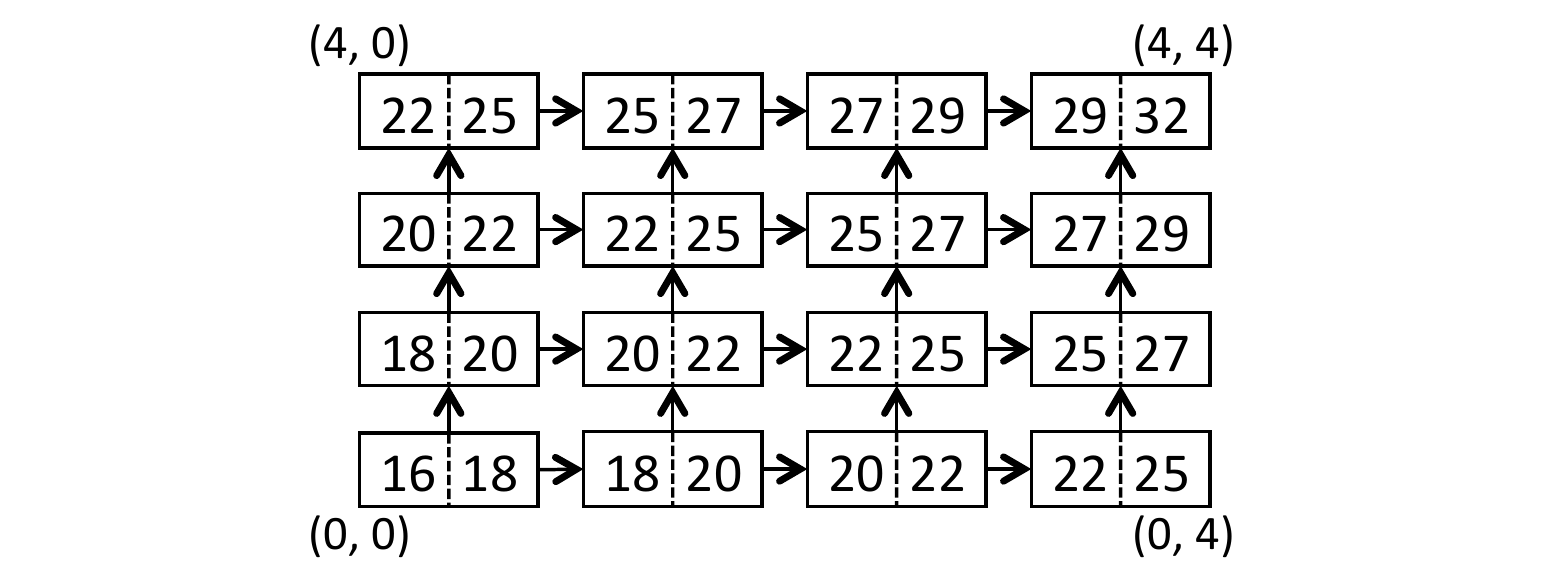}
 \caption{Number of channels of convolutional processing units in a two-dimensional grid layer ($N$=2, $L$=4), where the number of unit channels is $[16, 32]$.
Each box depicts a unit in a grid layer,
and a left number and a right number in the box are, respectively, the number of input channels and the number of output channels.}
 \label{fig:gridnet4}
\end{figure}

A grid layer contains numerous processing paths between input and output.
To improve the generalization abilities of the grid layer,
a feature map calculated using each processing path must be different from feature maps calculated using other paths.
For this purpose, in a SwGridNet, the number of channels of a convolutional processing unit differs from other units in the grid layer.
Figure~\ref{fig:gridnet4} presents an example of the number of unit channels in a two-dimensional grid layer.
The number of input channels and the number of output channels of a convolutional processing unit located at $(p_0,\cdots,p_{N-1})$ is defined as
\begin{eqnarray}
c^{\textrm{input}}_{p_0,\cdots,p_{N-1}} = \lfloor c^{\textrm{min}} + (c^{\textrm{max}} - c^{\textrm{min}})\frac{\sum p_i}{1+N(L-1)} \rfloor \\
c^{\textrm{output}}_{p_0,\cdots,p_{N-1}} = \lfloor c^{\textrm{min}} + (c^{\textrm{max}} - c^{\textrm{min}})\frac{1+\sum p_i}{1+N(L-1)} \rfloor
\end{eqnarray}
Here, $c^{\textrm{input}}_{p_0,\cdots,p_{N-1}}$ is the number of input channels of a unit at $(p_0,\cdots,p_{N-1})$;
$c^{\textrm{output}}_{p_0,\cdots,p_{N-1}}$ is also the number of output channels of the unit.
Parameters $c^{\textrm{min}}$ and $c^{\textrm{max}}$ respectively denote the minimum number of unit channels and the maximum number of unit channels.
The number of input channels of a unit equals the number of output channels of the neighbor units.
Therefore, channel modification such as zero-padding is not necessary for a SwGridNet.

The number of output channels of a split layer equals the sum of input channels of the convolutional processing units because all convolutional processing units receive their input matrices from the split layer.
Therefore, the number of output channels of a split layer is $\sum c^{\textrm{input}}_{p_0,\cdots,p_{N-1}}$.
In the same manner, the number of input channels of a join layer also equals the sum of output channels of the convolutional processing units because all convolutional processing units forward their output matrices to the join layer.
The number of input channels of a join layer is $\sum c^{\textrm{output}}_{p_0,\cdots,p_{N-1}}$

\section{Implementation and Experiment Results}
\label{sec:evaluation}

\subsection{Implementation}

\begin{figure}
\centering
 \includegraphics[width=\hsize]{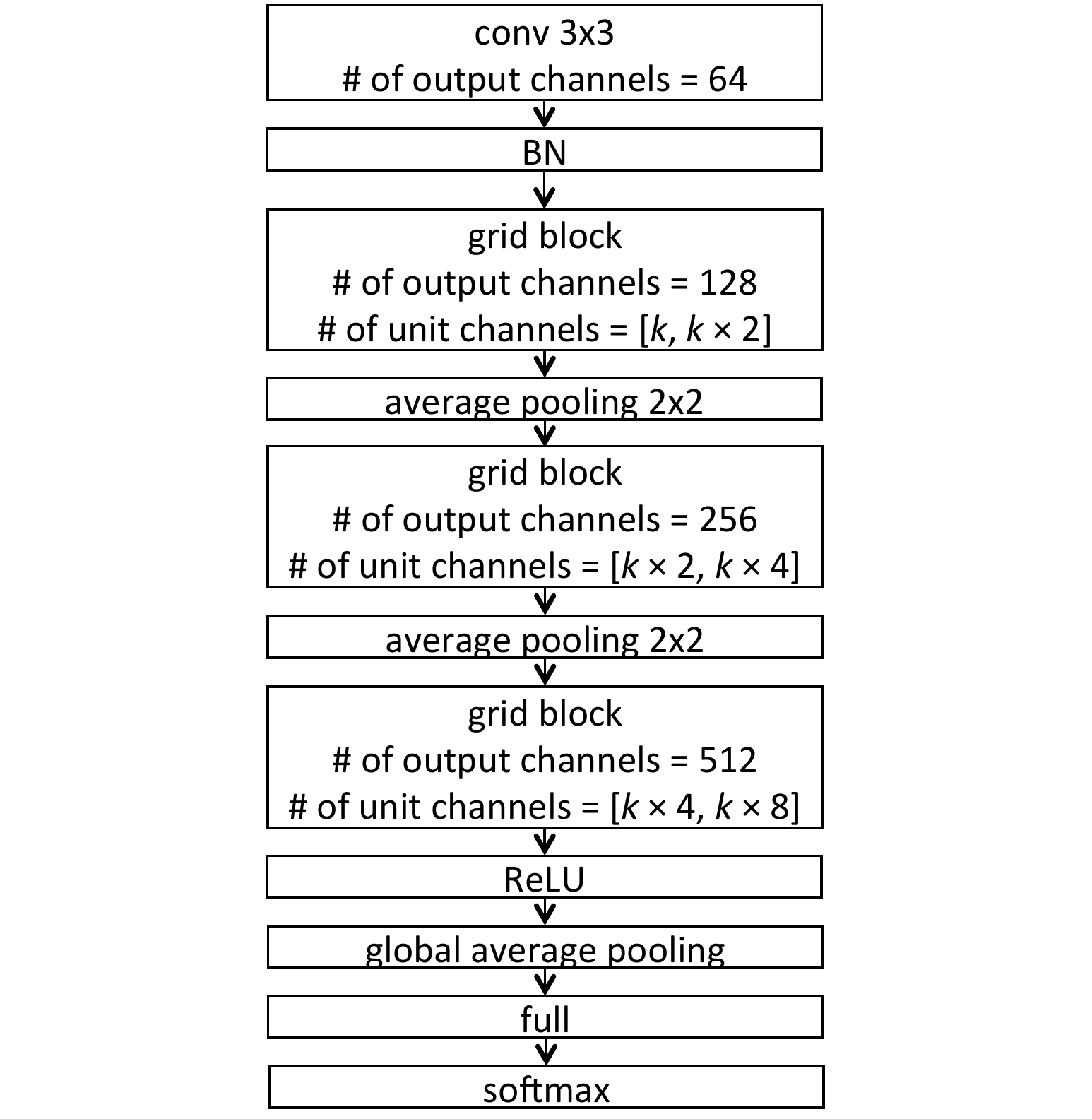}
 \caption{Network architecture of a SwGridNet for CIFAR-10 and CIFAR-100 classification tasks.}
 \label{fig:network}
\end{figure}

The author implemented SwGridNets\footnote{The repository is {\it https://github.com/takedarts/swgridnet}.} for CIFAR-10 and CIFAR-100 classification tasks using Chainer\footnote{the web site is {\it https://chainer.org}.}, which is a framework for neural networks.
The SwGridNets are implemented as a chain of grid blocks, as presented in Figure~\ref{fig:network}.
The SwGridNets increase the number of channels at each grid block and
decrease the resolution of feature maps after each grid block.
Unlike existing state-of-the-art deep CNNs~\cite{ResNet,PyramidalResNet,ResNeXt},
the SwGridNets use average pooling to decrease the resolution.
The basic concept of the SwGridNet implementations, however, is the same as the other state-of-the-art deep CNNs.

CIFAR-10 and CIFAR-100 are datasets used for image classification tasks~\cite{CIFAR}.
CIFAR-10 consists of 60,000 color images of 32 $\times$ 32 pixels, which are categorized into 10 classes.
Actually, CIFAR-100 includes the images, but the images are categorized into 100 classes.
For this study, 50,000 images were used in the datasets for training; 10,000 images were used in the datasets for validation.
For the training, parameters of the SwGridNets are initialized using MSRA~\cite{MSRAInit}.
The parameters are updated using Momentum SGD (momentum of 0.9, weight decay of 1e-4).
The SwGridNets for 630 epochs were trained with batch size of 128.
The learning rate is initially set to 0.2. Then it is changed according to Cosine Annealing with Warm Restart ($T_0$=10, $T_{\textrm{mult}}$=2)~\cite{WarmRestart}.
To improve the performance, I apply a standard data augmentation (random clipping and flip) to the images for training just like existing state-of-the-art deep CNNs~\cite{ResNet,PyramidalResNet,ResNeXt}.
Images for training, however, are not normalized using the mean or standard deviation.

\subsection{Dimensions of Grid Blocks}

\begin{table}
\centering
 \caption{Test error rates of SwGridNets in a CIFAR-10 classification task.
 The test error rates are median values of five runs.}
 \vskip 3pt
 \label{table:result1}
 \begin{tabular}{|l|c|c|}
  \hline
  settings & \# of params & error (\%) \\
  \hline
  $N$=1, $L$=16, $k$=16 & 3.7M & 4.56 \\
  $N$=2, $L$=4, $k$=16 & 3.7M & 4.39 \\
  $N$=4, $L$=2, $k$=16 & 3.8M & 4.36 \\
  \hline 
 \end{tabular}
\end{table}

\begin{table*}
\centering
 \caption{Test error rates of SwGridNets and state-of-the-art deep CNNs in CIFAR-10/100 classification tasks.
 The test error rates of SwGridNets are median values of five runs.}
 \vskip 3pt
 \label{table:result2}
 \begin{tabular}{|l|l|c|c|c|}
  \hline
  model & settings & \# of params & CIFAR-10 & CIFAR-100 \\
  \hline
  ResNet-110~\cite{ResNet} & & 1.7M & 6.43\% & \\
  \hline
  FractalNet-20~\cite{FractalNet} & with drop-path and dropout & 38.6M & 4.59\% & 23.36\% \\
  FractalNet-40 & & 22.9M & 5.21\% & 21.49\% \\
  \hline
  WideResNet-28~\cite{WideResNet} & $k$=10, with dropout & 36.5M & 3.89\% & 18.85\% \\
  \hline
  DenseNet-100~\cite{DenseNet} & $k$=24 & 27.2M & 3.74\% & 19.25\% \\
  & $k$=40, bottleneck and compression & 25.6M & 3.46\% & 17.18\% \\
  \hline
  PyramidNet-110~\cite{PyramidalResNet} & $\alpha$=84 & 3.8M & 4.26\% & 20.66\% \\
  & $\alpha$=270 & 28.3M & 3.73\% & 18.25\% \\
  PyramidNet-164 & $\alpha$=270, bottleneck & 27.0M & 3.48\% & 17.01\% \\
  \hline
  ResNeXt-29~\cite{ResNeXt} & 8$\times$64d & 34.4M & 3.65\% & 17.77\% \\
  & 16$\times$64d & 68.1M & 3.58\% & 17.31\% \\
  \hline 
  Shake-Shake-26~\cite{ShakeNet} & 2$\times$32 & 2.9M & 3.55\% & \\
  & 2$\times$96 & 26.2M & 2.86\% & \\
  Shake-Even-29 & 2$\times$4$\times$64 & 34.4M &  & 15.85\% \\
  \hline 
  Shake-Shake-26 + cutout~\cite{Cutout} & 2$\times$96 & 26.2M & {\bf 2.56\%} & \\
  Shake-Even-29 + cutout & 2$\times$4$\times$64 & 34.4M & & {\bf 15.20\%} \\
  \hline 
  SwGridNet (proposal) & $N$=2, $L$=4, $k$=16 & 3.7M & 4.39\% & \\
  & $N$=2, $L$=5, $k$=32 & 18.1M & 3.55\% & 17.77\% \\
  & $N$=2, $L$=5, $k$=32, ensemble of 4 models& 18.1M$\times$4 & 2.95\% & 15.67\% \\
  \hline 
 \end{tabular} \\ 
\end{table*}

Table~\ref{table:result1} presents the performance of SwGridNets for a CIFAR-10 classification task.
Here, $N$ and $L $ respectively denote the grid block and the side length dimensions of a of a grid block, as described in \ref{subsec:arch}.
In addition, $k$ denotes the minimum number of unit channels in a first grid block as presented in Figure~\ref{fig:network}.
The dimensions of grid blocks in these SwGridNets differ,
but these SwGridNets include the same number of convolutional processing units.

The performance of a four-dimensional SwGridNet ($N$=4, $L$=2) is better than that of a one-dimensional SwGridNet ($N$=1, $L$=16).
As described in \ref{subsec:path}, a four-dimensional SwGridNet includes more processing paths than a one-dimensional SwGridNet.
Therefore, a four-dimensional SwGridNet has a higher generalization ability than a one-dimensional SwGridNet.
Because these processing paths have the same effect as ensemble learning,
a four-dimensional SwGridNet obtains a lower error rate in the image classification task than a one-dimensional SwGridNet.

\subsection{Comparison with State-of-the-art Deep CNNs}

Table~\ref{table:result2} presents the respective performance parameters of SwGridNets and state-of-the-art deep CNNs in CIFAR-10 and CIFAR-100 classification tasks.
In this experiment, five models of SwGridNets are trained. Four models in the trained models are used to build ensemble models of SwGridNets.
The ensemble models of SwGridNets achieve a test error rate of 2.95\% in the CIFAR-10 classification task
and a test error rate of 15.67\% in the CIFAR-100 classification task.
The performance results of the SwGridNets are nearly equal to the performance results obtained for state-of-the-art deep CNNs,
even though the SwGridNets do not use popular generalization techniques such as Dropout~\cite{Dropout}, Stochastic Depth~\cite{StochasticResNet}, Shake Network~\cite{ShakeNet} or Cutout~\cite{Cutout}.
Because it is possible to apply the generalization techniques to a SwGridNet,
A SwGridNet with generalization techniques is expected to achieve better performance than current results presented in Table~\ref{table:result2}.

\section{Conclusion}
\label{sec:conclusion}

A SwGridNet described in this paper is a neural network in which many convolutional processing units are mutually connected as a grid network.
The grid network of convolutional processing units has many processing paths between input and output, which perform calculations using individual parameters.
A SwGridNet has a high generalization capability because the multipath network architecture has the same effect as ensemble learning.
Experimental results presented in this paper show that SwGridNets respectively achieve test error rates of 2.95\% and 15.67\% in a CIFAR-10 and CIFAR-100 classification tasks.
The experimentally obtained results demonstrate that the performance of SwGridNets is close to state-of-the-art deep CNNs.

{\small

}

\end{document}